\documentclass{article}

\usepackage{times}
\usepackage{graphicx} 
\usepackage{subfigure} 
\graphicspath{{figures/}}

\usepackage{natbib}

\usepackage{algorithm}
\usepackage{algorithmic}
\usepackage{amsmath}
\usepackage{amsfonts}
\usepackage{hyperref}
\usepackage{color}
\usepackage{soul}
\usepackage{multicol}

\usepackage[accepted]{icml2015}

\icmltitlerunning{Massively Parallel Methods for Deep Reinforcement Learning}

\begin{document} 

\twocolumn[
\icmltitle{Massively Parallel Methods for Deep Reinforcement Learning}

\icmlauthor{Arun Nair, Praveen Srinivasan, Sam Blackwell, Cagdas Alcicek, Rory Fearon, Alessandro De Maria, Vedavyas Panneershelvam, Mustafa Suleyman, Charles Beattie, Stig Petersen, Shane Legg, Volodymyr Mnih, Koray Kavukcuoglu, David Silver}
{  \\ \{arunsnair, prav, blackwells, cagdasalcicek, roryf, ademaria, darthveda, mustafasul, cbeattie, svp, legg, vmnih, korayk, davidsilver @google.com \}}
\icmladdress{Google DeepMind, London}

\icmlkeywords{reinforcement learning, deep learning}

\vskip 0.3in
]

\newcommand{\argmax}[1]{\underset{#1}{\operatorname{argmax}}\;}
\newcommand{\argmin}[1]{\underset{#1}{\operatorname{argmin}}\;}
\renewcommand{\max}[1]{\underset{#1}{\operatorname{max}}\;}
\renewcommand{\min}[1]{\underset{#1}{\operatorname{min}}\;}
\newcommand{\expect}[2]{\mathbb{E}_{#1} \left[ #2 \right]}
\newcommand{\TODO}[1]{\hl{TODO: #1}}

\begin{abstract} 
We present the first massively distributed architecture for deep reinforcement learning. This architecture uses four main components: parallel actors that generate new behaviour; parallel learners that are trained from stored experience; a distributed neural network to represent the value function or behaviour policy; and a distributed store of experience. We used our architecture to implement the Deep Q-Network algorithm (DQN)~\cite{mnih2013atari}. Our distributed algorithm was applied to 49 games from Atari 2600 games from the Arcade Learning Environment, using identical hyperparameters. Our performance surpassed non-distributed DQN in 41 of the 49 games and also reduced the wall-time required to achieve these results by an order of magnitude on most games.
\end{abstract}

\section{Introduction}
\label{submission}

Deep learning methods have recently achieved state-of-the-art results in vision and speech domains~\cite{krizhevsky-imagenet,simonyan-deep,szegedy-deep,graves2013speech,dahl2012context}, mainly due to their ability to automatically learn high-level features from a supervised signal. Recent advances in reinforcement learning (RL) have successfully combined deep learning with value function approximation, by using a deep convolutional neural network to represent the action-value (Q) function~\cite{mnih2013atari}. Specifically, a new method for training such deep Q-networks, known as DQN, has enabled RL to learn control policies in complex environments with high dimensional images as inputs \cite{mnih-dqn-2015}. This method outperformed a human professional in many games on the Atari 2600 platform, using the same network architecture and hyper-parameters. However, DQN has only previously been applied to single-machine architectures, in practice leading to long training times. For example, it took 12-14 days on a GPU to train the DQN algorithm on a single Atari game \cite{mnih-dqn-2015}. In this work, our goal is to build a distributed architecture that enables us to scale up deep reinforcement learning algorithms such as DQN by exploiting massive computational resources.

One of the main advantages of deep learning is that computation can be easily parallelized. In order to exploit this scalability, deep learning algorithms have made extensive use of hardware advances such as GPUs. However, recent approaches have focused on massively distributed architectures that can learn from more data in parallel and therefore outperform training on a single machine~\cite{coates2013deep,dean2012distbelief}. For example, the DistBelief framework \cite{dean2012distbelief} distributes the neural network parameters across many machines, and parallelizes the training by using  asynchronous stochastic gradient descent (ASGD). DistBelief has been used to achieve state-of-the-art results in several domains \cite{szegedy-deep} and has been shown to be much faster than single GPU training~\cite{dean2012distbelief}.

Existing work on distributed deep learning has focused exclusively on supervised and unsupervised learning. In this paper we develop a new architecture for the reinforcement learning paradigm. This architecture consists of four main components: parallel actors that generate new behaviour; parallel learners that are trained from stored experience; a distributed neural network to represent the value function or behaviour policy; and a distributed experience replay memory. 

A unique property of RL is that an agent influences the training data distribution by interacting with its environment. In order to generate more data, we deploy multiple agents running in parallel that interact with multiple instances of the same environment. Each such \emph{actor} can store its own record of past experience, effectively providing a distributed \emph{experience replay memory} with vastly increased capacity compared to a single machine implementation. Alternatively this experience can be explicitly aggregated into a distributed database. In addition to generating more data, distributed actors can explore the state space more effectively, as each actor behaves according to a slightly different policy. 

A conceptually distinct set of distributed \emph{learners} reads samples of stored experience from the experience replay memory, and updates the value function or policy according to a given RL algorithm. Specifically, we focus in this paper on a variant of the DQN algorithm, which applies ASGD updates to the parameters of the Q-network. As in DistBelief, the parameters of the Q-network may also be distributed over many machines.

We applied our distributed framework for RL, known as \emph{Gorila} (General Reinforcement Learning Architecture), to create a massively distributed version of the DQN algorithm. We applied Gorila DQN to 49 games on the Atari 2600 platform. We outperformed single GPU DQN on 41 games and outperformed human professional on 25 games. Gorila DQN also trained much faster than the non-distributed version in terms of wall-time, reaching the performance of single GPU DQN roughly ten times faster for most games.

\section{Related Work}

There have been several previous approaches to parallel or distributed RL. A significant part of this work has focused on distributed multi-agent systems \cite{weiss:distributedRL,lauer:distributedRL}. In this approach, there are many agents taking actions within a single shared environment, working cooperatively to achieve a common objective. While computation is distributed in the sense of decentralized control, these algorithms focus on effective teamwork and emergent group behaviors. Another paradigm which has been explored is concurrent reinforcement learning \cite{silver:concurrentRL}, in which an agent can interact in parallel with an inherently distributed environment, e.g. to optimize interactions with multiple users on the internet. Our goal is quite different to both these distributed and concurrent RL paradigms: we simply seek to solve a single-agent problem more efficiently by exploiting parallel computation. 

The MapReduce framework has been applied to standard MDP solution methods such as policy evaluation, policy iteration and value iteration, by distributing the computation involved in large matrix multiplications \cite{li:mapreduce}. However, this work is narrowly focused on batch methods for linear function approximation, and is not immediately applicable to non-linear representations using online reinforcement learning in environments with unknown dynamics.

Perhaps the closest prior work to our own is a parallelization of the canonical \emph{Sarsa} algorithm over multiple machines. Each machine has its own instance of the agent and environment \cite{grounds:parallelRL}, running a simple reinforcement learning algorithm (linear Sarsa, in this case). The changes to the parameters of the linear function approximator are periodically communicated using a peer-to-peer mechanism, focusing especially on those parameters that have changed most. In contrast, our architecture allows for client-server communication and a separation between acting, learning and parameter updates; furthermore we exploit much richer function approximators using a distributed framework for deep learning.

\section{Background}
\subsection{DistBelief}
DistBelief~\cite{dean2012distbelief} is a distributed system for training large neural networks on massive amounts of data efficiently by using two types of parallelism.
Model parallelism, where different machines are responsible for storing and training different parts of the model, is used to allow efficient training of models much larger than what is feasible on a single machine or GPU.
Data parallelism, where multiple copies or replicas of each model are trained on different parts of the data in parallel, allows for more efficient training on massive datasets than a single process.
We briefly discuss the two main components of the DistBelief architecture -- the central parameter server and the model replicas.

The central parameter server holds the master copy of the model. The job of the parameter server is to apply the incoming gradients from the replicas to the model and, when requested, to send its latest copy of the model to the replicas. The parameter server can be sharded across many machines and different shards apply gradients independently of other shards.

Each replica maintains a copy of the model being trained. This copy could be sharded across multiple machines if, for example, the model is too big to fit on a single machine. The job of the replicas is to calculate the gradients given a mini-batch, send them to the parameter server, and to periodically query the parameter server for an updated version of the model. The replicas send gradients and request updated parameters independently of each other and hence may not be synced to the same parameters at any given time.
\subsection{Reinforcement Learning}

\begin{figure}[ht]
    \begin{center}
        \centerline{\includegraphics[width=\columnwidth]{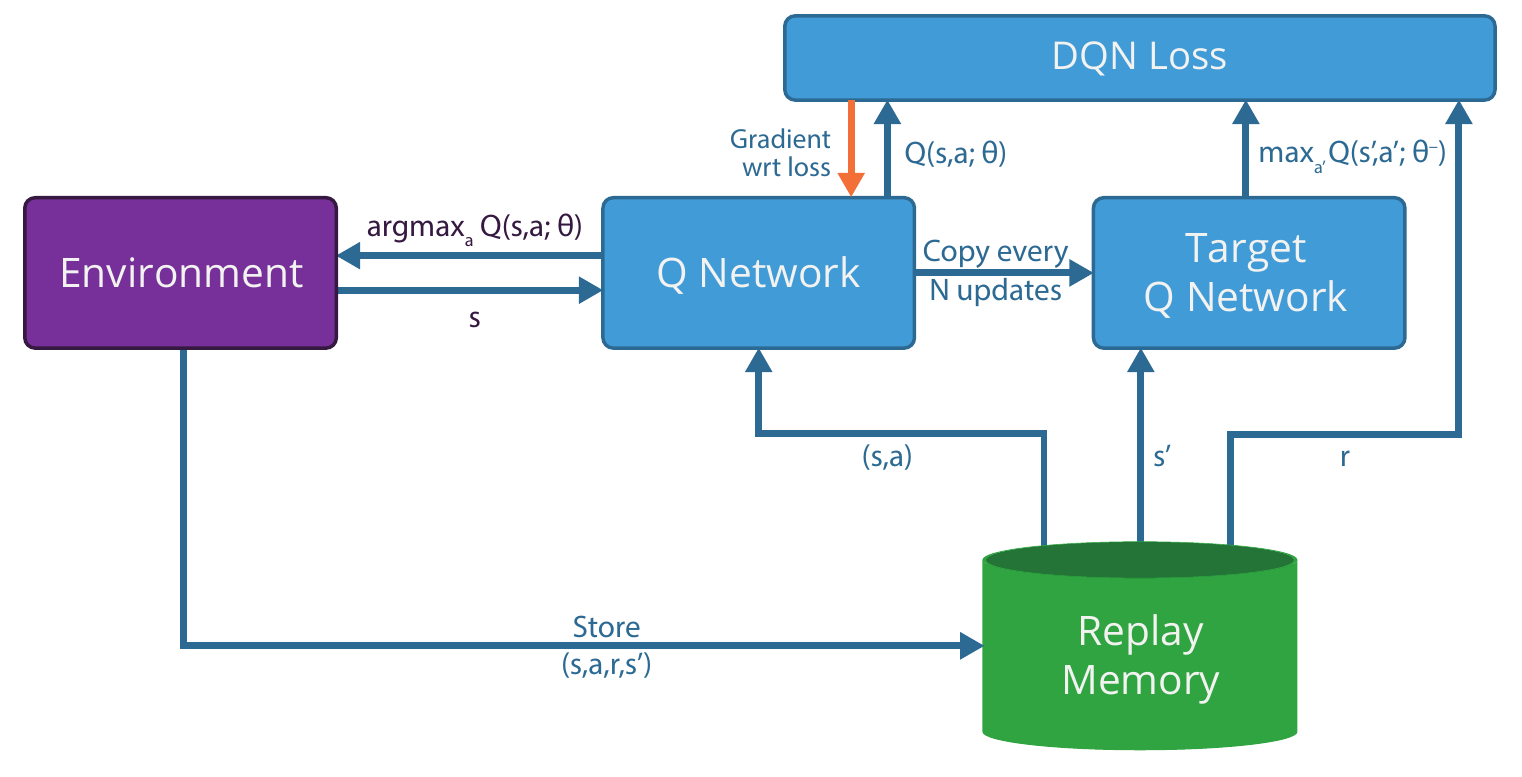}}
        \caption{The DQN algorithm is composed of three main components, the Q-network $(Q(s,a;\theta))$ that defines the behavior policy, the target Q-network $(Q(s,a;\theta^-))$ that is used to generate target Q values for the DQN loss term and the replay memory that the agent uses to sample random transitions for training the Q-network.}
        \label{DQN-figure}
    \end{center}
    \vskip -0.2in
\end{figure}

In the reinforcement learning (RL) paradigm, the agent interacts sequentially with an environment, with the goal of maximising cumulative rewards. At each step $t$ the agent observes state $s_t$, selects an action $a_t$, and receives a reward $r_t$. The agent's \emph{policy} $\pi(a | s)$ maps states to actions and defines its behavior. The goal of an RL agent is to maximize its expected total reward, where the rewards are discounted by a factor $\gamma \in [0,1]$ per time-step. Specifically, the \emph{return} at time $t$ is $R_t =\sum\limits_{t'=t}^{T} \gamma^{t'-t}r_{t'}$  where $T$ is the step when the episode terminates. The \emph{action-value function} $Q^\pi(s,a)$ is the expected return after observing state $s_t$ and taking an action under $a$ policy $\pi$, $Q^\pi(s,a) = \expect{}{R_t | s_t=s, a_t=a, \pi}$, and the \emph{optimal} action-value function is the maximum possible value that can be achieved by any policy, $Q^*(s,a) = \argmax{\pi}{Q^\pi(s,a)}$. The action-value function obeys a fundamental recursion known as the Bellman equation, $Q^*(s,a) = \expect{}{r + \gamma \ \max{a'}{Q^*(s',a')}}$. 

One of the core ideas behind reinforcement learning is to represent the action-value function using a function approximator such as a neural network, $Q(s,a) = Q(s,a; \theta)$. The parameters $\theta$ of the so-called \emph{Q-network} are optimized so as to approximately solve the Bellman equation. For example, the Q-learning algorithm iteratively updates the action-value function $Q(s,a; \theta)$ towards a sample of the Bellman target, $r + \gamma \ \max{a'}{Q(s',a'; \theta)}$. However, it is well-known that the Q-learning algorithm is highly unstable when combined with non-linear function approximators such as deep neural networks~\cite{tsitsiklis-convergence}. 

\subsection{Deep Q-Networks}
Recently, a new RL algorithm has been developed which is in practice much more stable when combined with deep Q-networks \cite{mnih2013atari,mnih-dqn-2015}. Like Q-learning, it iteratively solves the Bellman equation by adjusting the parameters of the Q-network towards the Bellman target. However, DQN, as shown in Figure~\ref{DQN-figure} differs from Q-learning in two ways. First, DQN uses experience replay \cite{lin1993reinforcement}. At each time-step $t$ during an agent's interaction with the environment it stores the experience tuple $e_t = (s_t, a_t, r_t, s_{t+1})$ into a replay memory $D_t = \{e_1, ..., e_t\}$. 

Second, DQN maintains two separate Q-networks $Q(s,a; \theta)$ and $Q(s,a; \theta^-)$ with current parameters $\theta$ and old parameters $\theta^-$ respectively. The current parameters $\theta$ may be updated many times per time-step, and are copied into the old parameters $\theta^-$ after $N$ iterations. At every update iteration $i$ the current parameters $\theta$ are updated so as to minimise the mean-squared Bellman error with respect to old parameters $\theta^-$, by optimizing the following loss function (DQN Loss),
\begin{small}
\begin{align}
L_i(\theta_i) = \expect{}{\left( r + \gamma \ \max{a'} Q(s', a'; \theta_i^-) - Q(s, a; \theta_i) \right)^2}
\end{align}
\end{small}
For each update $i$, a tuple of experience $(s,a,r,s') \sim U(D)$ (or a minibatch of such samples) is sampled uniformly from the replay memory $D$. For each sample (or minibatch), the current parameters $\theta$ are updated by a stochastic gradient descent algorithm. Specifically, $\theta$ is adjusted in the direction of the sample gradient $g_i$ of the loss with respect to $\theta$,
\begin{small}
\begin{align}
g_i = \left( r + \gamma \ \max{a'} Q(s', a'; \theta_i^-) - Q(s, a; \theta_i) \right) \nabla_{\theta_i} Q(s,a; \theta)
\label{eqn:dqngrad}
\end{align}
\end{small}
Finally, actions are selected at each time-step $t$ by an $\epsilon$-greedy behavior with respect to the current Q-network $Q(s,a;\theta)$.

\section{Distributed Architecture}

\begin{figure*}[ht]
	\begin{center}
		\centerline{\includegraphics[width=\textwidth]{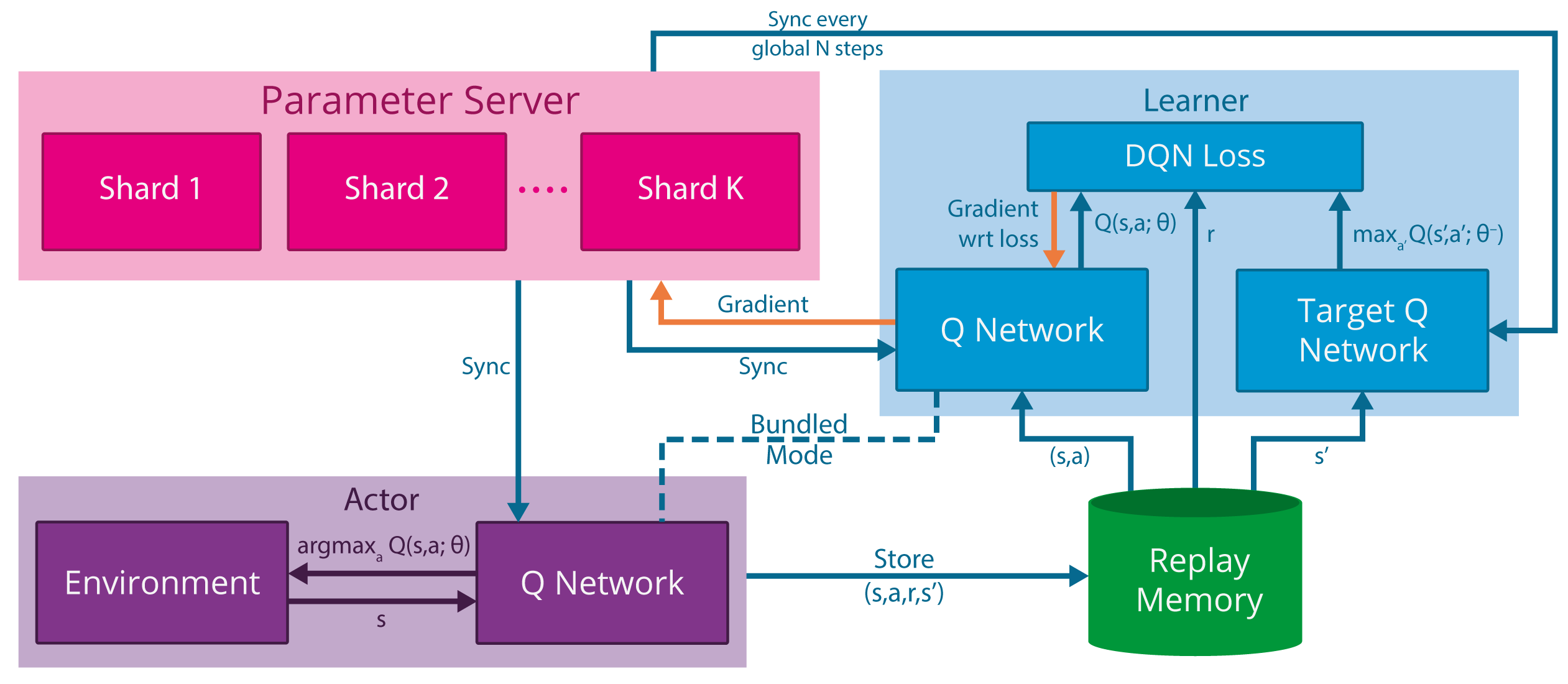}}
		\caption{The Gorila agent parallelises the training procedure by separating out learners, actors and parameter server. In a single experiment, several learner processes exist and they continuously send the gradients to parameter server and receive updated parameters. At the same time, independent actors can also in parallel accumulate experience and update their Q-networks from the parameter server. }
		\label{Gorila-figure}
	\end{center}
	\vskip -0.2in
\end{figure*}

\begin{algorithm}[t]
	\caption{Distributed DQN Algorithm}
	\label{alg:DistDQNAlgo}
	\begin{algorithmic}
		\STATE Initialise replay memory $D$ to size $P$.
		\STATE Initialise the training network for the action-value function $Q(s,a;\theta)$ with weights $\theta$ and target network $Q(s,a;\theta^-)$ with weights $\theta^- = \theta$.
		\FOR{$episode=1$ {\bfseries to} $M$}
		\STATE Initialise the start state to $s_{1}$.
		\STATE Update $\theta$ from parameters $\theta^+$ of the parameter server.
		\FOR{$t=1$ {\bfseries to} $T$}
		\STATE With probability $\epsilon$ take a random action $a_{t}$ or else $a_{t}$ = $\argmax{a}{Q(s, a ;\theta)}$.
		\STATE Execute the action in the environment and observe the reward $r_t$ and the next state $s_{t+1}$. Store $(s_t,a_t,r_t,s_{t+1})$ in $D$.
		\STATE Update $\theta$ from parameters $\theta^+$ of the parameter server.
		\STATE Sample random mini-batch from $D$. And for each tuple $(s_i,a_i,r_i,s_{i+1})$ set target $y_t$ as
		\IF{$s_{i+1} $ is $terminal$} 
		\STATE $y_t = r_i$
		\ELSE
		\STATE $y_t = r_i + \gamma \max{a'}{Q(s_{i+1}, a' ; \theta^{-})}$
		\ENDIF
		\STATE Calculate the loss $L_t = (y_t - Q(s_i, a_i ; \theta)^2)$.
		\STATE Compute gradients with respect to the network parameters $\theta$ using equation~\ref{eqn:dqngrad}.
		\STATE Send gradients to the parameter server.
		\STATE Every global $N$ steps sync $\theta^{-}$ with parameters $\theta^+$ from the parameter server.
		\ENDFOR
		\ENDFOR
	\end{algorithmic}
\end{algorithm}

We now introduce \emph{Gorila} (General Reinforcement Learning Architecture), a framework for massively distributed reinforcement learning. The Gorila architecture, shown in Figure~\ref{Gorila-figure} contains the following components:

{\bf Actors}. Any reinforcement learning agent must ultimately select actions $a_t$ to apply in its environment. We refer to this process as \emph{acting}. The Gorila architecture contains $N_{act}$ different actor processes, applied to $N_{act}$ corresponding instantiations of the same environment. Each actor $i$ generates its own trajectories of experience $s^i_1, a^i_1, r^i_1, ..., s^i_T, a^i_T, r^i_T$ within the environment, and as a result each actor may visit different parts of the state space. The quantity of experience that is generated by the actors after $T$ time-steps is approximately $TN_{act}$. Each actor contains a replica of the Q-network, which is used to determine behavior, for example using an $\epsilon$-greedy policy. The parameters of the Q-network are synchronized periodically from the parameter server.

{\bf Experience replay memory}. The experience tuples $e^i_t = (s^i_t, a^i_t, r^i_t, s^i_{t+1})$ generated by the actors are stored in a replay memory $D$. We consider two forms of experience replay memory. First, a \emph{local} replay memory stores each actor's experience $D^i_t =  \{ e^i_1, ..., e^i_t \}$ locally on that actor's machine. If a single machine has sufficient memory to store $M$ experience tuples, then the overall memory capacity becomes $MN_{act}$. Second, a \emph{global} replay memory aggregates the experience into a distributed database. In this approach the overall memory capacity is independent of $N_{act}$ and may be scaled as desired, at the cost of additional communication overhead.

{\bf Learners}. Gorila contains $N_{learn}$ learner processes. Each learner contains a replica of the Q-network and its job is to compute desired changes to the parameters of the Q-network. For each learner update $k$, a minibatch of experience tuples $e = (s,a,r,s')$ is sampled from either a local or global experience replay memory $D$ (see above). The learner applies an off-policy RL algorithm such as DQN \cite{mnih2013atari} to this minibatch of experience, in order to generate a gradient vector $g_i$.\footnote{The experience in the replay memory is generated by old behavior policies which are most likely different to the current behavior of the agent; therefore all updates must be performed off-policy \cite{sutton:book}.} The gradients $g_i$ are communicated to the parameter server; and the parameters of the Q-network are updated periodically from the parameter server. 

{\bf Parameter server}. Like DistBelief, the Gorila architecture uses a central parameter server to maintain a distributed representation of the Q-network $Q(s,a; \theta^+)$. The parameter vector $\theta^+$ is split disjointly across $N_{param}$ different machines. Each machine is responsible for applying gradient updates to a subset of the parameters. The parameter server receives gradients from the learners, and applies these gradients to modify the parameter vector $\theta^+$, using an asynchronous stochastic gradient descent algorithm. 

The Gorila architecture provides considerable flexibility in the number of ways an RL agent may be parallelized. It is possible to have parallel acting to generate large quantities of data into a global replay database, and then process that data with a single serial learner. In contrast, it is possible to have a single actor generating data into a local replay memory, and then have multiple learners process this data in parallel to learn as effectively as possible from this experience. However, to avoid any individual component from becoming a bottleneck, the Gorila architecture in general allows for arbitrary numbers of actors, learners, and parameter servers to both generate data, learn from that data, and update the model in a scalable and fully distributed fashion.

The simplest overall instantiation of Gorila, which we consider in our subsequent experiments, is the \emph{bundled} mode in which there is a one-to-one correspondence between actors, replay memory, and learners ($N_{act} = N_{learn}$). Each bundle has an actor generating experience, a local replay memory to store that experience, and a learner that updates parameters based on samples of experience from the local replay memory. The only communication between bundles is via parameters: the learners communicate their gradients to the parameter server; and the Q-networks in the actors and learners are periodically synchronized to the parameter server.

\subsection{Gorila DQN}

We now consider a specific instantiation of the Gorila architecture implementing the DQN algorithm.
As described in the previous section, the DQN algorithm utilizes two copies of the Q-network: a current Q-network with parameters $\theta$ and a target Q-network with parameters $\theta^-$.
The DQN algorithm is extended to the distributed implementation in Gorila as follows.
The parameter server maintains the current parameters $\theta^+$ and the actors and learners contain replicas of the current Q-network $Q(s,a;\theta)$ that are synchronized from the parameter server before every acting step.

The learner additionally maintains the target Q-network $Q(s,a;\theta^-)$.
The learner's target network is updated from the parameter server $\theta^+$ after every $N$ gradient updates in the central parameter server.

Note that $N$ is a global parameter that counts the total number of updates to the central parameter server rather than counting the updates from the local learner.

The learners generate gradients using the DQN gradient given in Equation~\ref{eqn:dqngrad}. However, the gradients are not applied directly, but instead communicated to the parameter server. The parameter server then applies the updates that are accumulated from many learners.

\subsection{Stability}
While the DQN training algorithm was designed to ensure stability of training neural networks with reinforcement learning, training using a large cluster of machines running multiple other tasks poses additional challenges.
The Gorila DQN implementation uses additional safeguards to ensure stability in the presence of disappearing nodes, slowdowns in network traffic, and slowdowns of individual machines.
One such safeguard is a parameter that determines the maximum time delay between the local parameters $\theta$ (the gradients $g_i$ are computed using $\theta$) and the parameters $\theta^+$ in the parameter server.

All gradients older than the threshold are discarded by the parameter server.  Additionally, each actor/learner keeps a running average and standard deviation of the absolute DQN loss for the data it sees and discards gradients with absolute loss higher than the mean plus several standard deviations.  Finally, we used the AdaGrad update rule ~\cite{duchi2011adaptive}.

\section{Experiments}
\subsection{Experimental Set Up}
We evaluated Gorila by conducting experiments on 49 Atari 2600 games using the Arcade Learning Environment~\cite{bellemare-ale}.
Atari games provide a challenging and diverse set of reinforcement learning problems where an agent must learn to play the games directly from $210\times 160$ RGB video input with only the changes in the score provided as rewards.
We closely followed the experimental setup of DQN~\cite{mnih-dqn-2015} using the same preprocessing and network architecture.
We preprocessed the $210\times 160$ RGB images by downsampling them to $84\times 84$ and extracting the luminance channel.

The Q-network $Q(s,a;\theta)$ had 3 convolutional layers followed by a fully-connected hidden layer.
The $84\times 84\times 4$ input to the network is obtained by concatenating the images from four previous preprocessed frames.
The first convolutional layer had $32$ filters of size $4\times 8\times 8$ and stride $4$. The second convolutional layer had $64$ filters of size $32\times 4\times 4$ with stride $2$, while the third had $64$ filters with size $64\times 3\times 3$ and stride $1$. The next layer had $512$ fully-connected output units, which is followed by a linear fully-connected output layer with a single output unit for each valid action. 
Each hidden layer was followed by a rectifier nonlinearity.

We have used the same frame skipping step implemented in~\cite{mnih-dqn-2015} by repeating every action $a_t$ over the next $4$ frames. 

In all experiments, Gorila DQN used: $N_{param}=31$ and $N_{learn} = N_{act} =$ 100. We use the bundled mode. Replay memory size $D=$ 1 million frames and used $\epsilon$-greedy as the behaviour policy with $\epsilon$ annealed from 1 to 0.1 over the first one million global updates.
Each learner syncs the parameters $\theta^-$ of its target network after every 60K parameter updates performed in the parameter server.

\begin{figure*}[!ht]
	\vskip -0.1in
	\begin{center}
		\centerline{\includegraphics[width=0.95\textwidth]{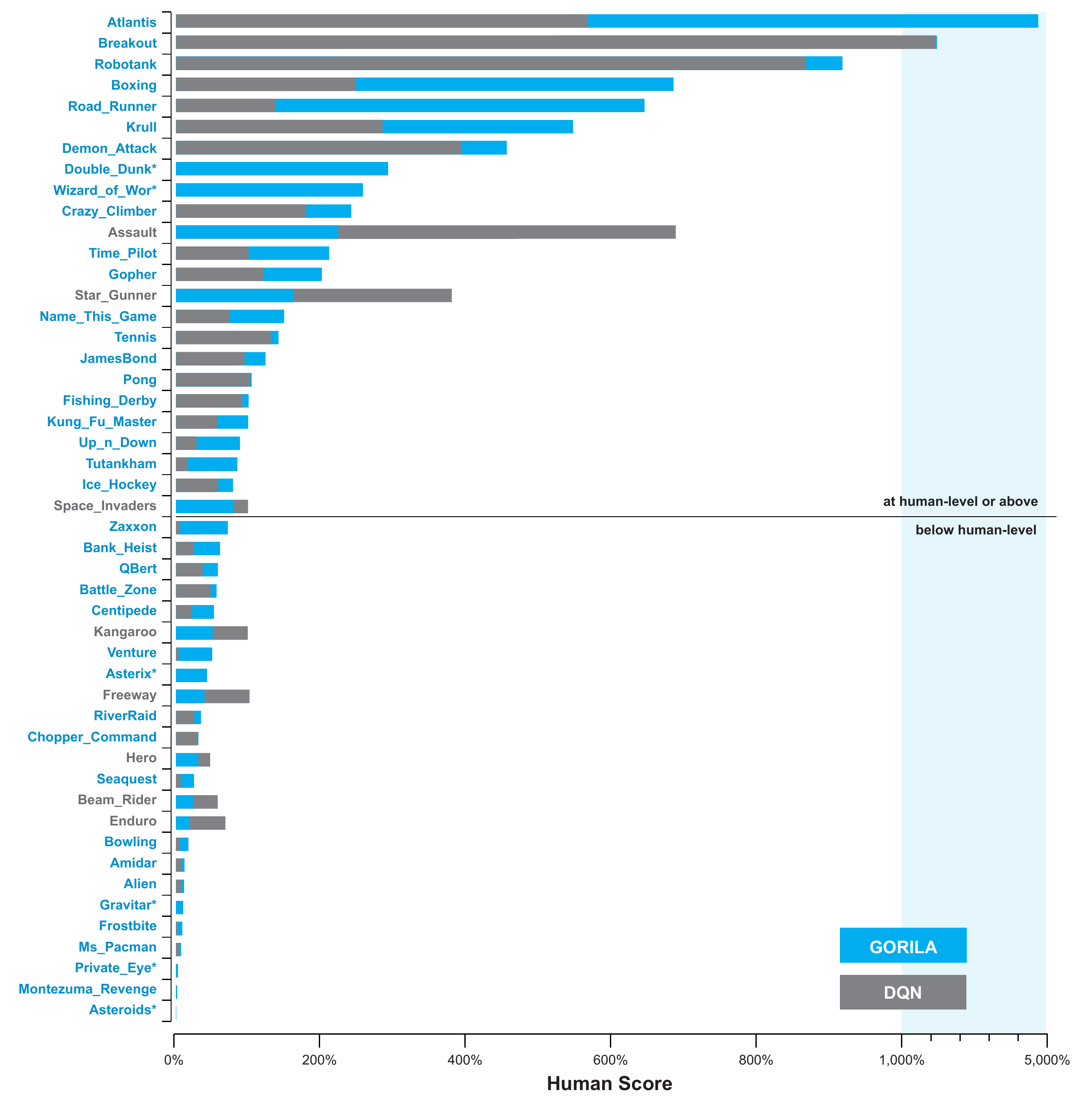}}
		\caption{Performance of the Gorila agent on 49 Atari games with human starts evaluation compared with DQN~\cite{mnih-dqn-2015} performance with scores normalized to expert human performance.  Font color indicates which method has the higher score. *Not showing DQN scores for Asterix, Asteroids, Double Dunk, Private Eye, Wizard Of Wor and Gravitar because the DQN human starts scores are less than the random agent baselines. Also not showing Video Pinball because the human expert scores are less than the random agent scores.}
		\label{fig:humanstarts}
	\end{center}
	\vskip -0.2in
\end{figure*}

\begin{figure*} [!ht]
	\vskip -0.1in
	\begin{center}
		\centerline{\includegraphics[width=0.95\textwidth]{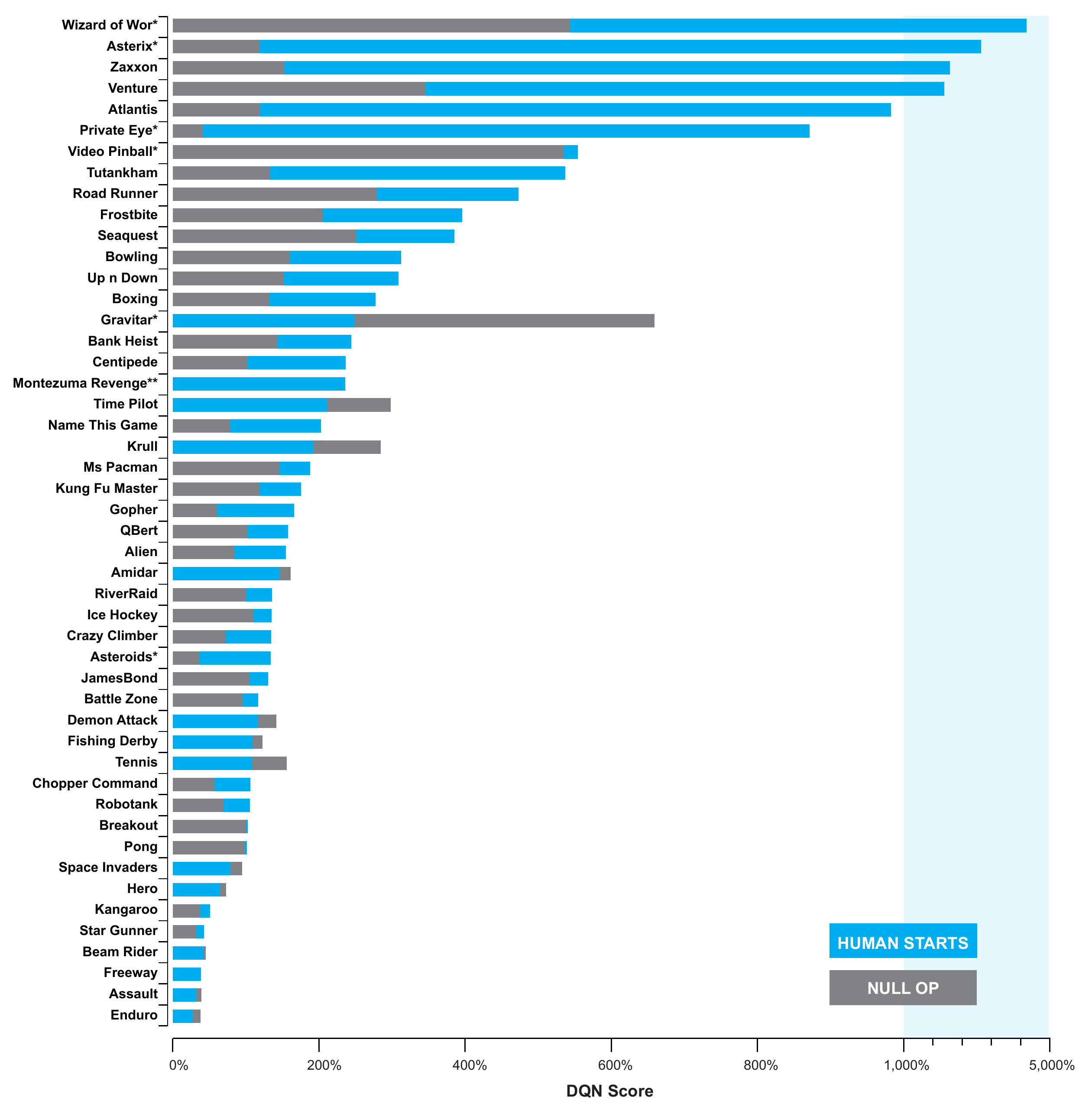}}
		\caption{Performance of the Gorila agent on 49 Atari games with human starts and null op evaluations normalized with respect to DQN human start and null op scores respectively. This figure shows the generalization improvements of Gorila compared to DQN. *Using a score of 0 for the human starts random agent score for Asterix, Asteroids, Double Dunk, Private Eye, Wizard Of Wor and Gravitar because the human starts DQN scores are less than the random agent scores. Not showing Double Dunk because both the DQN scores and the random agent scores are negative. **Not showing null op scores for Montezuma Revenge because both the human start scores and random agent scores are 0.
		}
		\label{fig:humanstarts1}
	\end{center}
	\vskip -0.2in
\end{figure*}

\subsection{Evaluation}

We used two types of evaluations.
The first follows the protocol established by DQN. Each trained agent was evaluated on 30 episodes of the game it was trained on.
A random  number of frames were skipped by repeatedly taking the null or do nothing action before giving control to the agent in order to ensure variation in the initial conditions.
The agents were allowed to play until the end of the game or up to 18000 frames (5 minutes), whichever came first, and the scores were averaged over all 30 episodes. We refer to this evaluation procedure as {\it null op starts}.

Testing how well an agent generalizes is especially important in the Atari domain because the emulator is completely deterministic.

Our second evaluation method, which we call {\it human starts}, aims to measure how well the agent generalizes to states it may not have trained on. To that end, we have introduced 100 random starting points that were sampled from a human professional's gameplay for each game.
To evaluate an agent, we ran it from each of the 100 starting points until the end of the game or until a total of 108000 frames (equivalent to 30 minutes) were played counting the frames the human played to reach the starting point.
The total score accumulated only by the agent (not considering any points won by the human player) were averaged to obtain the evaluation score.

In order to make it easier to compare results on 49 games with a greatly varying range of scores we present the results on a scale where 0 is the score obtained by a random agent and 100 is the score obtained by a professional human game player.
The random agent selected actions uniformly at random at 10Hz and it was evaluated using the same starting states as the agents for both kinds of evaluations ({\it null op} starts and {\it human starts}).

We selected hyperparameter values by performing an informal search on the games of Breakout, Pong and Seaquest which were then fixed for all the games.
We have trained Gorila DQN 5 times on each game using the same fixed hyperparameter settings and random network initializations. Following DQN, we periodically evaluated each model during training and kept the best performing network parameters for the final evaluation. We average these final evaluations over the 5 runs, and compare the mean evaluations with DQN and human expert scores.


\section {Results}

\begin{figure}[ht]
	\vskip -0.1in
	\begin{center}
		\centerline{\includegraphics[width=\columnwidth]{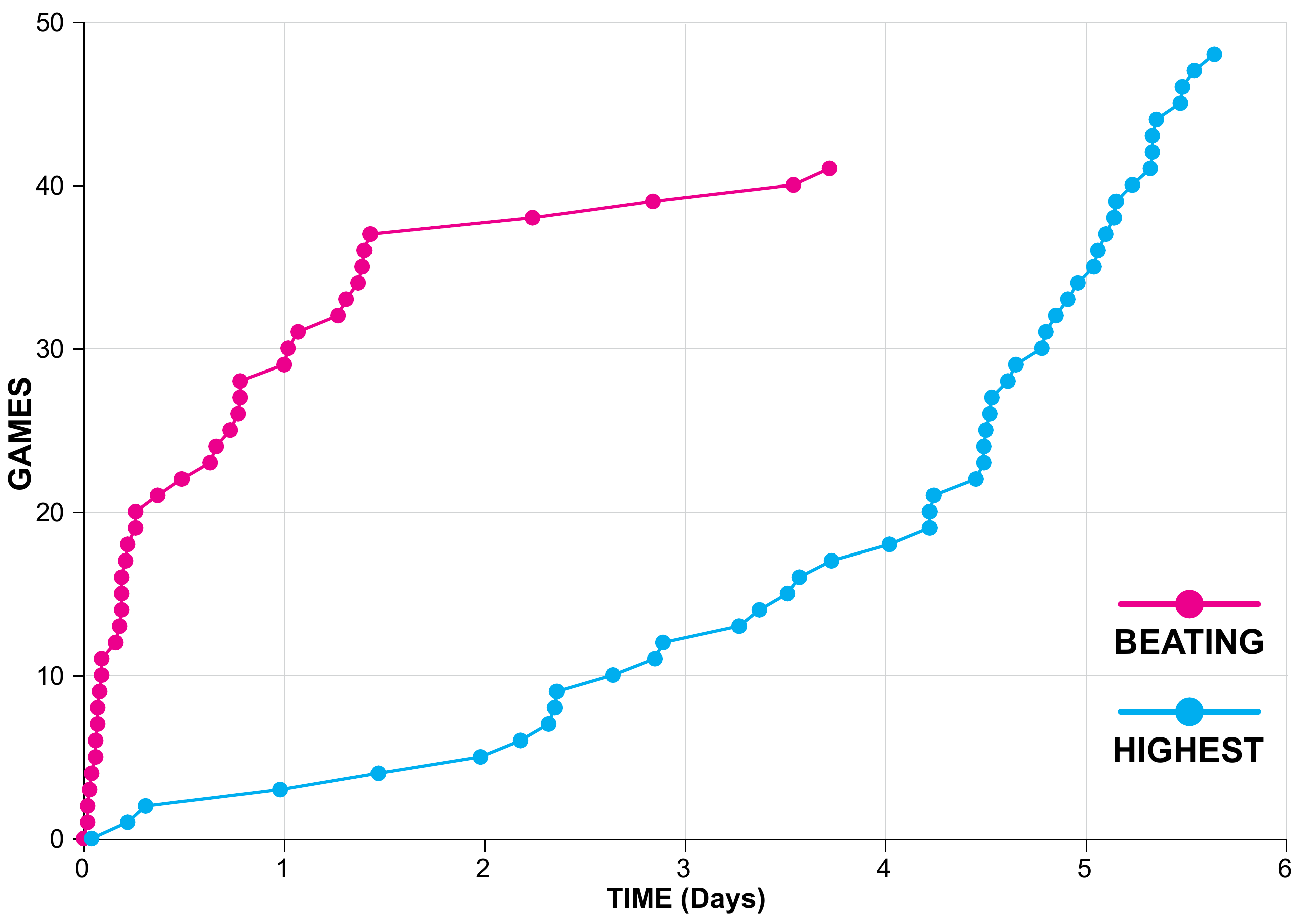}}
		\caption{
			The time required by Gorila DQN to surpass single DQN performance (red curve) and to reach its peak performance (blue curve). 
		}
		\label{fig:time}
	\end{center}
	\vskip -0.2in
\end{figure}

We first compared Gorila DQN agents trained for up to 6 days to single GPU DQN agents trained for 12-14 days.
Figure~\ref{fig:humanstarts} shows the normalized scores under the human starts evaluation.
Using human starts Gorila DQN outperformed single GPU DQN on 41 out of 49 games given roughly one half of the training time of single GPU DQN.
On 22 of the games Gorila DQN obtained double the score of single GPU DQN, and on 11 games Gorila DQN's score was 5 times higher.
Similarly, using the original {\it null op starts} evaluation  Gorila DQN outperformed the single GPU DQN on 31 out of 49 games. 
These results show that parallel training significantly improved performance in less training time. Also, better results on {\it human starts} compared to {\it null op starts} suggest that Gorila DQN is especially good at generalizing to potentially unseen states compared to single GPU DQN.  Figure~\ref{fig:humanstarts1} further illustrates these improvements in generalization by showing Gorila DQN scores with human starts normalized with respect to GPU DQN scores with human starts (blue bars) and Gorila DQN scores from null op starts normalized by GPU DQN scores from null op starts (gray bars).
In fact, Gorila DQN performs at a level similar or superior to a human professional (75\% of the human score or above) in 25 games despite starting from states sampled from human play.
One possible reason for the improved generalization is the significant increase in the number of states Gorila DQN sees by using 100 parallel actors.

We next look at how the performance of Gorila DQN improved during training.
Figure~\ref{fig:time} shows how quickly Gorila DQN reached the performance of single GPU DQN and how quickly Gorila DQN reached its own best score under the human starts evaluation.
Gorila DQN surpassed the best single GPU DQN scores on 19 games in 6 hours, 23 games in 12 hours, 30 in 24 hours and 38 games in 36 hours (red curve).
This is a roughly an order of magnitude reduction in training time required to reach the single process DQN score.
On some games Gorila DQN achieved its best score in under two days but for most of the games the performance keeps improving with longer training time (blue curve).

\section{Conclusion}
In this paper we have introduced the first massively distributed architecture for deep reinforcement learning. The \emph{Gorila} architecture acts and learns in parallel, using a distributed replay memory and distributed neural network. We applied Gorila to  an asynchronous variant of the state-of-the-art DQN algorithm. A single machine had previously achieved state-of-the-art results in the challenging suite of Atari 2600 games, but it was not previously known whether the good performance of DQN would continue to scale with additional computation. By leveraging massive parallelism, Gorila DQN significantly outperformed single-GPU DQN on 41 out of 49 games; achieving by far the best results in this domain to date. Gorila takes a further step towards fulfilling the promise of deep learning in RL: a scalable architecture that performs better and better with increased computation and memory.

\bibliographystyle{icml2015}
\bibliography{gorilabib}
\graphicspath{{figures/}}

	\title{Appendix}
	\maketitle

\section{Appendix}

\subsection{Data}
We present all the data that has been used in the paper. 
	\begin{itemize}
		\item Table 1 shows the various normalized scores for null op evaluation.
		\item Table 2 shows the various normalized scores for human start evaluation.
		\item Table 3 shows the various raw scores for human start evaluation.
		\item Table 4 shows the various raw scores for null op evaluation.
	\end{itemize}

\begin{table*}[h]
	\vskip 0.5in
	\caption {NULL OP NORMALIZED}
	\centering 
	\begin{tabular}{l|r|r|r}
		\hline \hline
		Games & DQN & Gorila & Gorila \\
		&  (human normalized) & (human normalized) & (DQN normalized) \\
		\hline\
		Alien & 42.74 & 35.99 & 84.20 \\
		\hline\
		Amidar & 43.93 & 70.89 & 161.36 \\
		\hline\
		Assault & 246.16 & 96.39 & 39.15 \\
		\hline\
		Asterix & 69.95 & 75.04 & 107.26 \\
		\hline\
		Asteroids & 7.31 & 2.64 & 36.09 \\
		\hline\
		Atlantis & 451.84 & 539.11 & 119.31 \\
		\hline\
		Bank\_Heist & 57.69 & 82.58 & 143.15 \\
		\hline\
		Battle\_Zone & 67.55 & 64.63 & 95.68 \\
		\hline\
		Beam\_Rider & 119.79 & 54.31 & 45.34 \\
		\hline\
		Bowling & 14.65 & 23.47 & 160.18 \\
		\hline\
		Boxing & 1707.14 & 2256.66 & 132.18 \\
		\hline\
		Breakout & 1327.24 & 1330.56 & 100.25 \\
		\hline\
		Centipede & 62.98 & 64.23 & 101.97 \\
		\hline\
		Chopper\_Command & 64.77 & 37.00 & 57.12 \\
		\hline\
		Crazy\_Climber & 419.49 & 305.06 & 72.72 \\
		\hline\
		Demon\_Attack & 294.19 & 416.74 & 141.65 \\
		\hline\
		Double\_Dunk & 16.12 & 257.34 & 1595.55 \\
		\hline\
		Enduro & 97.48 & 37.11 & 38.07 \\
		\hline\
		Fishing\_Derby & 93.51 & 115.11 & 123.09 \\
		\hline\
		Freeway & 102.36 & 39.49 & 38.58 \\
		\hline\
		Frostbite & 6.16 & 12.64 & 205.23 \\
		\hline\
		Gopher & 400.42 & 243.35 & 60.77 \\
		\hline\
		Gravitar & 5.35 & 35.27 & 659.37 \\
		\hline\
		Hero & 76.50 & 56.14 & 73.38 \\
		\hline\
		Ice\_Hockey & 79.33 & 87.49 & 110.27 \\
		\hline\
		JamesBond & 145.00 & 152.50 & 105.16 \\
		\hline\
		Kangaroo & 224.20 & 83.71 & 37.33 \\
		\hline\
		Krull & 277.01 & 788.85 & 284.76 \\
		\hline\
		Kung\_Fu\_Master & 102.37 & 121.38 & 118.57 \\
		\hline\
		Montezuma\_Revenge & 0.00 & 0.09 & 0.00 \\
		\hline\
		Ms\_Pacman & 13.02 & 19.01 & 146.03 \\
		\hline\
		Name\_This\_Game & 278.28 & 218.05 & 78.35 \\
		\hline\
		Pong & 132 & 130 & 98.48 \\
		\hline\
		Private\_Eye & 2.53 & 1.04 & 41.05 \\
		\hline\
		QBert & 78.48 & 80.14 & 102.10 \\
		\hline\
		RiverRaid & 57.30 & 57.54 & 100.41 \\
		\hline\
		Road\_Runner & 232.91 & 651.00 & 279.50 \\
		\hline\
		Robotank & 509.27 & 352.92 & 69.29 \\
		\hline\
		Seaquest & 25.94 & 65.13 & 251.08 \\
		\hline\
		Space\_Invaders & 121.48 & 115.36 & 94.96 \\
		\hline\
		Star\_Gunner & 598.08 & 192.79 & 32.23 \\
		\hline\
		Tennis & 148.99 & 232.70 & 156.18 \\
		\hline\
		Time\_Pilot & 100.92 & 300.86 & 298.11 \\
		\hline\
		Tutankham & 112.22 & 149.53 & 133.24 \\
		\hline\
		Up\_n\_Down & 92.68 & 140.70 & 151.81 \\
		\hline\
		Venture & 32.00 & 104.87 & 327.71 \\
		\hline\
		Video\_Pinball & 2539.36 & 13576.75 & 534.65 \\
		\hline\
		Wizard\_of\_Wor & 67.48 & 314.04 & 465.32 \\
		\hline\
		Zaxxon & 54.08 & 77.63 & 143.53 \\
		\hline
	\end{tabular}
\end{table*}

\begin{table*}[h]
	\vskip 0.5in
	\caption{HUMAN STARTS NORMALIZED}
	\centering 
	\begin{tabular}{l|r|r|r}
		\hline \hline
		Games & DQN & Gorila & Gorila \\
		& (human normalized)&(human normalized)& (DQN normalized) \\
		\hline
		Alien & 7.07 & 10.97 & 155.06 \\
		\hline
		Amidar & 7.95 & 11.60 & 145.85 \\
		\hline
		Assault & 685.15 & 222.71 & 32.50 \\
		\hline
		Asterix & -0.54 & 42.87 & 2670.44 \\
		\hline
		Asteroids & -0.50 & 0.15 & 133.93 \\
		\hline
		Atlantis & 477.76 & 4695.72 & 982.84 \\
		\hline
		Bank\_Heist & 24.82 & 60.64 & 244.32 \\
		\hline
		Battle\_Zone & 47.50 & 55.57 & 116.98 \\
		\hline
		Beam\_Rider & 57.23 & 24.25 & 42.38 \\
		\hline
		Bowling & 5.39 & 16.85 & 312.62 \\
		\hline
		Boxing & 245.94 & 682.03 & 277.31 \\
		\hline
		Breakout & 1149.42 & 1184.15 & 103.02 \\
		\hline
		Centipede & 22.00 & 52.06 & 236.59 \\
		\hline
		Chopper\_Command & 28.98 & 30.74 & 106.06 \\
		\hline
		Crazy\_Climber & 178.54 & 240.52 & 134.71 \\
		\hline
		Demon\_Attack & 390.38 & 453.60 & 116.19 \\
		\hline
		Double\_Dunk & -350.00 & 290.62 & 0.00 \\
		\hline
		Enduro & 67.81 & 18.59 & 27.42 \\
		\hline
		Fishing\_Derby & 90.99 & 99.44 & 109.28 \\
		\hline
		Freeway & 100.78 & 39.23 & 38.92 \\
		\hline
		Frostbite & 2.19 & 8.70 & 395.82 \\
		\hline
		Gopher & 120.41 & 200.05 & 166.13 \\
		\hline
		Gravitar & -1.01 & 10.20 & 248.67 \\
		\hline
		Hero & 46.87 & 30.43 & 64.92 \\
		\hline
		Ice\_Hockey & 57.84 & 78.23 & 135.25 \\
		\hline
		JamesBond & 94.02 & 122.53 & 130.31 \\
		\hline
		Kangaroo & 98.37 & 50.43 & 51.27 \\
		\hline
		Krull & 283.33 & 544.42 & 192.14 \\
		\hline
		Kung\_Fu\_Master & 56.49 & 99.18 & 175.57 \\
		\hline
		Montezuma\_Revenge & 0.60 & 1.41 & 236.00 \\
		\hline
		Ms\_Pacman & 3.72 & 7.01 & 188.30 \\
		\hline
		Name\_This\_Game & 73.13 & 148.38 & 202.88 \\
		\hline
		Pong & 102.08 & 103.63 & 101.51 \\
		\hline
		Private\_Eye & -0.57 & 3.04 & 871.41 \\
		\hline
		QBert & 36.55 & 57.71 & 157.89 \\
		\hline
		RiverRaid & 25.20 & 34.23 & 135.80 \\
		\hline
		Road\_Runner & 135.72 & 642.10 & 473.07 \\
		\hline
		Robotank & 863.07 & 913.69 & 105.86 \\
		\hline
		Seaquest & 6.41 & 24.69 & 385.13 \\
		\hline
		Space\_Invaders & 98.81 & 78.03 & 78.97 \\
		\hline
		Star\_Gunner & 378.03 & 161.04 & 42.60 \\
		\hline
		Tennis & 129.93 & 140.84 & 108.39 \\
		\hline
		Time\_Pilot & 99.57 & 210.13 & 211.01 \\
		\hline
		Tutankham & 15.68 & 84.19 & 536.80 \\
		\hline
		Up\_n\_Down & 28.33 & 87.50 & 308.76 \\
		\hline
		Venture & 3.52 & 49.50 & 1403.88 \\
		\hline
		Video\_Pinball & -4.65 & 1904.86 & 554.14 \\
		\hline
		Wizard\_of\_Wor & -14.87 & 256.58 & 4240.24 \\
		\hline
		Zaxxon & 4.46 & 71.34 & 1596.74 \\
		\hline
	\end{tabular}
\end{table*}

\begin{table*}[h]
	\vskip 0.5in
	\caption{RAW DATA - HUMAN STARTS}
	\centering 
	\begin{tabular}{l|r|r|r|r}
		\hline \hline
		Games & Random & Human  & DQN & Gorila Avg \\
		\hline
		Alien & 128.30 & 6371.30 & 570.20 & 813.54 \\
		\hline
		Amidar & 11.80 & 1540.40 & 133.40 & 189.15 \\
		\hline
		Assault & 166.90 & 628.90 & 3332.30 & 1195.85 \\
		\hline
		Asterix & 164.50 & 7536.00 & 124.50 & 3324.70 \\
		\hline
		Asteroids & 877.10 & 36517.30 & 697.10 & 933.63 \\
		\hline
		Atlantis & 13463.00 & 26575.00 & 76108.00 & 629166.50 \\
		\hline
		Bank\_Heist & 21.70 & 644.50 & 176.30 & 399.42 \\
		\hline
		Battle\_Zone & 3560.00 & 33030.00 & 17560.00 & 19938.00\\
		\hline
		Beam\_Rider & 254.60 & 14961.00 & 8672.40 & 3822.07 \\
		\hline
		Bowling & 35.20 & 146.50 & 41.20 & 53.95 \\
		\hline
		Boxing & -1.50 & 9.60 & 25.80 & 74.20 \\
		\hline
		Breakout & 1.60 & 27.90 & 303.90 & 313.03 \\
		\hline
		Centipede & 1925.50 & 10321.90 & 3773.10 & 6296.87 \\
		\hline
		Chopper\_Command & 644.00 & 8930.00 & 3046.00 & 3191.75 \\
		\hline
		Crazy\_Climber & 9337.00 & 32667.00 & 50992.00 & 65451.00 \\
		\hline
		Demon\_Attack & 208.30 & 3442.80 & 12835.20 & 14880.13 \\
		\hline
		Double\_Dunk & -16.00 & -14.40 & -21.60 & -11.35 \\
		\hline
		Enduro & -81.80 & 740.20 & 475.60 & 71.04 \\
		\hline
		Fishing\_Derby & -77.10 & 5.10 & -2.30 & 4.64 \\
		\hline
		Freeway & 0.20 & 25.60 & 25.80 & 10.16 \\
		\hline
		Frostbite & 66.40 & 4202.80 & 157.40 & 426.60 \\
		\hline
		Gopher & 250.00 & 2311.00 & 2731.80 & 4373.04 \\
		\hline
		Gravitar & 245.50 & 3116.00 & 216.50 & 538.37 \\
		\hline
		Hero & 1580.30 & 25839.40 & 12952.50 & 8963.36 \\
		\hline
		Ice\_Hockey & -9.70 & 0.50 & -3.80 & -1.72 \\
		\hline
		JamesBond & 33.50 & 368.50 & 348.50 & 444.00 \\
		\hline
		Kangaroo & 100.00 & 2739.00 & 2696.00 & 1431.00 \\
		\hline
		Krull & 1151.90 & 2109.10 & 3864.00 & 6363.09 \\
		\hline
		Kung\_Fu\_Master & 304.00 & 20786.80 & 11875.00 & 20620.00 \\
		\hline
		Montezuma\_Revenge & 25.00 & 4182.00 & 50.00 & 84.00 \\
		\hline
		Ms\_Pacman & 197.80 & 15375.00 & 763.50 & 1263.05 \\
		\hline
		Name\_This\_Game & 1747.80 & 6796.00 & 5439.90 & 9238.50 \\
		\hline
		Pong & -18.00 & 15.50 & 16.20 & 16.71 \\
		\hline
		Private\_Eye & 662.80 & 64169.10 & 298.20 & 2598.55 \\
		\hline
		QBert & 271.80 & 12085.00 & 4589.80 & 7089.83 \\
		\hline
		RiverRaid & 588.30 & 14382.20 & 4065.30 & 5310.27 \\
		\hline
		Road\_Runner & 200.00 & 6878.00 & 9264.00 & 43079.80 \\
		\hline
		Robotank & 2.40 & 8.90 & 58.50 & 61.78 \\
		\hline
		Seaquest & 215.50 & 40425.80 & 2793.90 & 10145.85 \\
		\hline
		Space\_Invaders & 182.60 & 1464.90 & 1449.70 & 1183.29 \\
		\hline
		Star\_Gunner & 697.00 & 9528.00 & 34081.00 & 14919.25 \\
		\hline
		Tennis & -21.40 & -6.70 & -2.30 & -0.69 \\
		\hline
		Time\_Pilot & 3273.00 & 5650.00 & 5640.00 & 8267.80 \\
		\hline
		Tutankham & 12.70 & 138.30 & 32.40 & 118.45 \\
		\hline
		Up\_n\_Down & 707.20 & 9896.10 & 3311.30 & 8747.67 \\
		\hline
		Venture & 18.00 & 1039.00 & 54.00 & 523.40 \\
		\hline
		Video\_Pinball & 20452.00 & 15641.10 & 20228.10 & 112093.37 \\
		\hline
		Wizard\_of\_Wor & 804.00 & 4556.00 & 246.00 & 10431.00 \\
		\hline
		Zaxxon & 475.00 & 8443.00 & 831.00 & 6159.40 \\
		\hline
	\end{tabular}
\end{table*}

\begin{table*}[h]
	\vskip  0.5in
	\caption{RAW DATA - NULL OP}
	\centering 
	\begin{tabular}{l|r|r|r|r}
		\hline \hline
		Games & Random & Human & DQN & Gorila Avg \\
		\hline
		Alien & 227.80 & 6875.40 & 3069.30 & 2620.53 \\
		\hline
		Amidar & 5.80 & 1675.80 & 739.50 & 1189.70 \\
		\hline
		Assault & 222.40 & 1496.40 & 3358.60 & 1450.41 \\
		\hline
		Asterix & 210.00 & 8503.30 & 6011.70 & 6433.33 \\
		\hline
		Asteroids & 719.10 & 13156.70 & 1629.30 & 1047.66 \\
		\hline
		Atlantis & 12850.00 & 29028.10 & 85950.00 & 100069.16 \\
		\hline
		Bank\_Heist & 14.20 & 734.40 & 429.70 & 609.00 \\
		\hline
		Battle\_Zone & 2360.00 & 37800.00 & 26300.00 & 25266.66 \\
		\hline
		Beam\_Rider & 363.90 & 5774.70 & 6845.90 & 3302.91 \\
		\hline
		Bowling & 23.10 & 154.80 & 42.40 & 54.01 \\
		\hline
		Boxing & 0.10 & 4.30 & 71.80 & 94.88 \\
		\hline
		Breakout & 1.70 & 31.80 & 401.20 & 402.20 \\
		\hline
		Centipede & 2090.90 & 11963.20 & 8309.40 & 8432.30 \\
		\hline
		Chopper\_Command & 811.00 & 9881.80 & 6686.70 & 4167.50 \\
		\hline
		Crazy\_Climber & 10780.50 & 35410.50 & 114103.30 & 85919.16 \\
		\hline
		Demon\_Attack & 152.10 & 3401.30 & 9711.20 & 13693.12 \\
		\hline
		Double\_Dunk & -18.60 & -15.50 & -18.10 & -10.62 \\
		\hline
		Enduro & 0.00 & 309.60 & 301.80 & 114.90 \\
		\hline
		Fishing\_Derby & -91.70 & 5.50 & -0.80 & 20.19 \\
		\hline
		Freeway & 0.00 & 29.60 & 30.30 & 11.69 \\
		\hline
		Frostbite & 65.20 & 4334.70 & 328.30 & 605.16 \\
		\hline
		Gopher & 257.60 & 2321.00 & 8520.00 & 5279.00 \\
		\hline
		Gravitar & 173.00 & 2672.00 & 306.70 & 1054.58 \\
		\hline
		Hero & 1027.00 & 25762.50 & 19950.30 & 14913.87 \\
		\hline
		Ice\_Hockey & -11.20 & 0.90 & -1.60 & -0.61 \\
		\hline
		JamesBond & 29.00 & 406.70 & 576.70 & 605.00 \\
		\hline
		Kangaroo & 52.00 & 3035.00 & 6740.00 & 2549.16 \\
		\hline
		Krull & 1598.00 & 2394.60 & 3804.70 & 7882.00 \\
		\hline
		Kung\_Fu\_Master & 258.50 & 22736.20 & 23270.00 & 27543.33 \\
		\hline
		Montezuma\_Revenge & 0.00 & 4366.70 & 0.00 & 4.16 \\
		\hline
		Ms\_Pacman & 307.30 & 15693.40 & 2311.00 & 3233.50 \\
		\hline
		Name\_This\_Game & 2292.30 & 4076.20 & 7256.70 & 6182.16 \\
		\hline
		Pong & -20.70 & 9.30 & 18.90 & 18.30 \\
		\hline
		Private\_Eye & 24.90 & 69571.30 & 1787.60 & 748.60 \\
		\hline
		QBert & 163.90 & 13455.00 & 10595.80 & 10815.55 \\
		\hline
		RiverRaid & 1338.50 & 13513.30 & 8315.70 & 8344.83 \\
		\hline
		Road\_Runner & 11.50 & 7845.00 & 18256.70 & 51007.99 \\
		\hline
		Robotank & 2.20 & 11.90 & 51.60 & 36.43 \\
		\hline
		Seaquest & 68.40 & 20181.80 & 5286.00 & 13169.06 \\
		\hline
		Space\_Invaders & 148.00 & 1652.30 & 1975.50 & 1883.41 \\
		\hline
		Star\_Gunner & 664.00 & 10250.00 & 57996.70 & 19144.99 \\
		\hline
		Tennis & -23.80 & -8.90 & -1.60 & 10.87 \\
		\hline
		Time\_Pilot & 3568.00 & 5925.00 & 5946.70 & 10659.33 \\
		\hline
		Tutankham & 11.40 & 167.60 & 186.70 & 244.97 \\
		\hline
		Up\_n\_Down & 533.40 & 9082.00 & 8456.30 & 12561.58 \\
		\hline
		Venture & 0.00 & 1187.50 & 380.00 & 1245.33 \\
		\hline
		Video\_Pinball & 16256.90 & 17297.60 & 42684.10 & 157550.21 \\
		\hline
		Wizard\_of\_Wor & 563.50 & 4756.50 & 3393.30 & 13731.33 \\
		\hline
		Zaxxon & 32.50 & 9173.30 & 4976.70 & 7129.33 \\
		\hline
	\end{tabular}
\end{table*}

\end{document}